\documentclass{article}
\usepackage{spconf,amsmath,graphicx}

\usepackage{float}
\usepackage{subfig}
\usepackage{multirow}
\usepackage{color}
\usepackage[colorlinks, linkcolor=blue, anchorcolor=blue, citecolor=blue]
{hyperref}
\usepackage{marvosym}
\usepackage{ifsym}

\title{iSmallNet: Densely Nested Network with Label Decoupling for Infrared Small Target Detection}
\name{Zhiheng Hu$^1$, Yongzhen Wang$^1$, Peng Li$^1$, Jie Qin$^1 $\Letter\thanks{\Letter~Corresponding Author: Jie Qin (jie.qin@nuaa.edu.cn).}, Haoran Xie$^2$, Mingqiang Wei$^1$}
\address{$^1$Nanjing University of Aeronautics and Astronautics\\
$^2$Lingnan University}
\begin{document}
%
\maketitle
\begin{abstract}
 Small targets are often submerged in cluttered backgrounds of infrared images. Conventional detectors tend to generate false alarms, while CNN-based detectors lose small targets in deep layers. To this end, we propose iSmallNet, a multi-stream densely nested network with label decoupling for infrared small object detection. On the one hand, to fully exploit the shape information of small targets, we decouple the original labeled ground-truth (GT) map into an interior map and a boundary one. The GT map, in collaboration with the two additional maps, tackles the unbalanced distribution of small object boundaries. On the other hand, two key modules are delicately designed and incorporated into the proposed network to boost the overall performance. First, to maintain small targets in deep layers, we develop a multi-scale nested interaction module to explore a wide range of context information. Second, we develop an interior-boundary fusion module to integrate multi-granularity information. Experiments on NUAA-SIRST and NUDT-SIRST clearly show the superiority of iSmallNet over 11 state-of-the-art detectors.
\end{abstract}

\section{Introduction}
 Infrared imaging captures invisible infrared images that represent the thermal energy radiated from objects and then converts them into visible images. 
 It belongs to contactless and passive reconnaissance with a long-distance observation capability in all-day/weather conditions, showing its good adaptability for monitoring `small' objects. 
 Owing to the above, infrared small target detection (ISTD) has been receiving increasing attention, with a wide range of applications in military reconnaissance \cite{2010-Classification-TK, 2016-Small-DSLYZ}, precise guidance \cite{2021-Infrared-SYA}, autonomous driving \cite{2018-Nighttime-LSXZD}, and \emph{etc}.

 The goal of ISTD is to locate small targets by generating their accurate 2D bounding boxes. 
 Although many efforts have been made, ISTD is still not well solved due to the non-trivial infrared imaging mechanism, where captured infrared images are of low contrast and low signal-to-noise ratio. 
 Consequently, various targets are very small and blurred, and even submerged in cluttered backgrounds.

 To detect small and weak targets in infrared images, traditional methods include filtering \cite{1996-Detection-RJ, 1999-Max-DMRC}, local contrast \cite{2013-A-CLWXT, 2014-A-HMYZBFFLKF, 2019-A-HMFLZ}, and low rank based approaches \cite{2013-Infrared-GMYWZH, 2017-Reweighted-DW, 2020-Infrared-ZLDLX}. 
 However, these algorithms rely heavily on hand-crafted features and fixed hyper-parameters, leading to inaccurate detection results in cluttered backgrounds. 
 Different from the traditional efforts, CNN-based methods \cite{2017-Image-LDZDHW} learn the characteristics of infrared small targets in a data-driven way.
 Most existing works \cite{2020-Infrared-MVM,2019-TBCNet-ZCYFLW,2020-Asymmetric-DWZB} have been developed upon general object detection/segmentation frameworks. 
 For example, McIntosh \textit{et al.} \cite{2020-Infrared-MVM} adapted several existing generic object detection networks for ISTD. 
 Dai \textit{et al.} \cite{2020-Asymmetric-DWZB} designed an asymmetric contextual module (ACM) to replace the plain skip connection of U-Net \cite{2015-U-RFB}. 
 Despite achieving promising results, most of them are not specifically designed for ISTD. 
 Since the size of infrared small targets is much smaller than general targets, applying these methods to ISTD can easily lead to the loss of small targets in deep layers.

 \begin{figure*}[!t]
	\centering
	\includegraphics[width=0.80\linewidth]{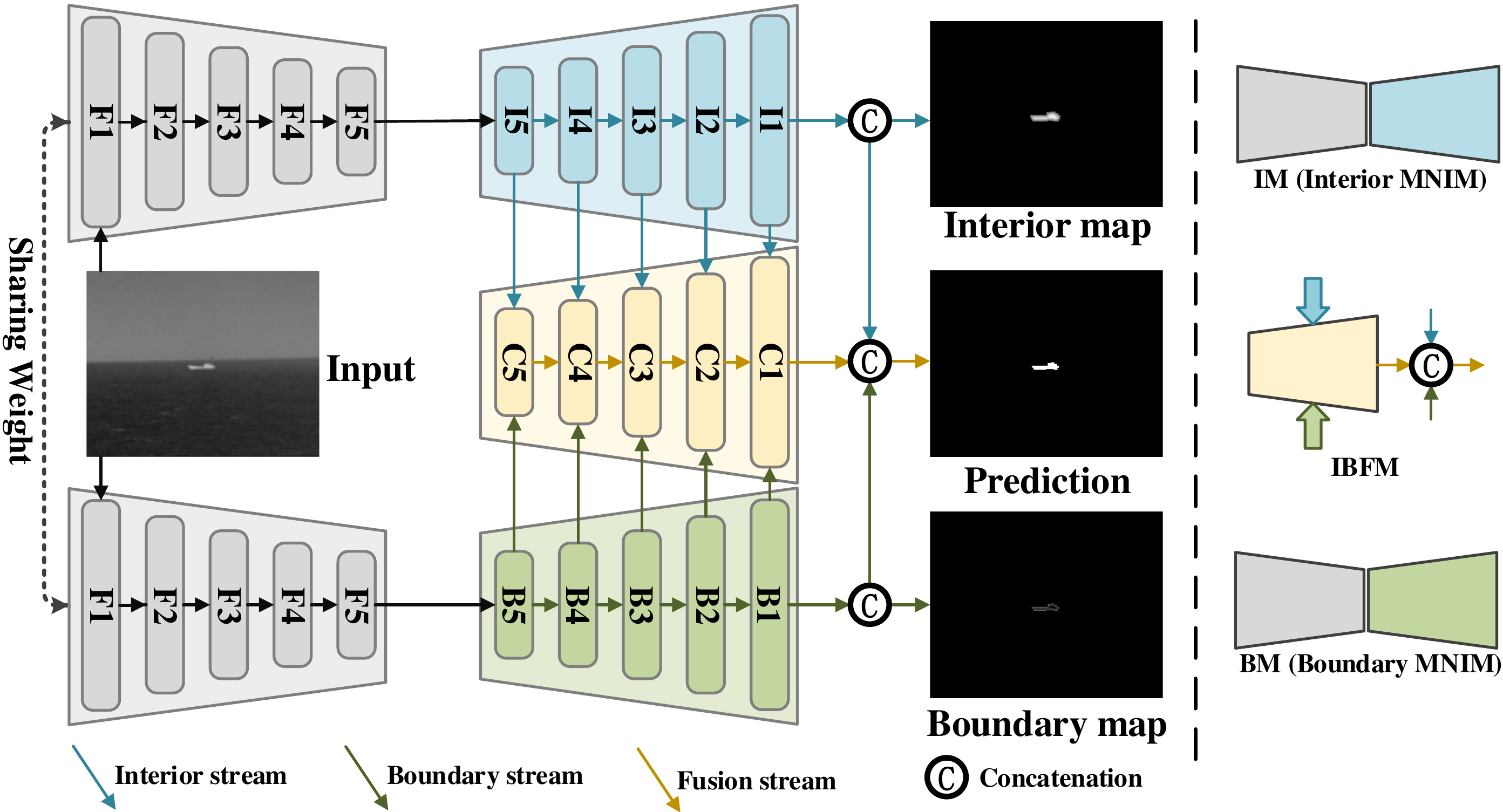}
	\caption{Overall framework of iSmallNet. iSmallNet consists of two encoders and three decoders, \emph{i.e.}, two backbone encoders for feature extraction, an interior decoder and a boundary decoder with multi-scale nested interaction module (MNIM) to produce interior/boundary maps, and a fusion decoder with interior-boundary fusion module (IBFM) for fusing multi-scale features.}
	\label{f:architecture} \vspace{-3mm}
 \end{figure*}
 
 Based on the above observations, we propose iSmallNet, a novel densely nested network with label decoupling for ISTD. 
 iSmallNet aims at accurately detecting small yet blurred targets submerged in cluttered backgrounds.
 Specifically, we propose a multi-stream architecture, which not only focuses on the entire target but also captures the essential cues from its interior and boundary. 
 To this end, label decoupling is employed to decompose the original map into two complementary ones. 
 Moreover, to fully unleash the potential information from the interior and boundary maps, we develop a multi-scale nested module to maintain the representations of small targets in very deep layers. 
 As a result, more accurate and robust detection results can be obtained, as shown in our experiments. 
 In summary, our main contributions include:
 \begin{itemize}
     \item We propose iSmallNet to fully exploit the potential cues from the interior and boundary of small targets. To the best of our knowledge, this kind of complementary information has never been explored in ISTD.
     \item To maintain the characteristics of small targets, we develop a multi-scale nested interaction module to explore a wide range of context information, as well as an interior-boundary fusion module to integrate the multi-granularity information.
     \item Experiments on two benchmarks show that iSmallNet outperforms 11 cutting-edge competitors and is more robust to the size and shape of small targets.
 \end{itemize}
 
\section{Methodology}
 iSmallNet is carefully designed based on the following observations. 
 Targets in infrared images are often small and vague. 
 The pixels belonging to the boundaries of small targets have a severely imbalanced distribution. 
 Hence, to enhance the detection ability, it is desirable to pay attention to both boundary and interior information of small targets. 
 To fulfill this, the original labeled GT map is decomposed into a boundary map and an interior map \cite{2020-Label-WWWST}. 
 On the one hand, the boundary map provides more boundary information around small targets; 
 on the other hand, the interior map concentrates on the center areas of small targets by discarding boundary information. 
 Although we can capture richer information by label decoupling, detailed information may still be lost in deep layers. 
 To this end, we propose a multi-scale nested interaction module (MNIM) to explore a wide range of context information, and also develop an interior-boundary fusion module (IBFM) to integrate multi-granularity information.

 \begin{figure*}[!t]
	\centering
	\includegraphics[width=1\linewidth]{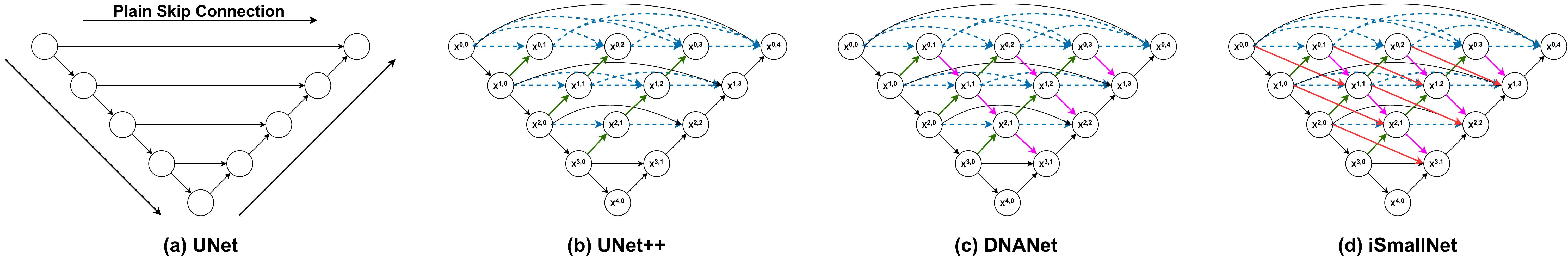}
	\caption{Illustration of three U-shape structures and our multi-scale nested interaction module.}
	\label{f:MNIM} \vspace{-3mm}
 \end{figure*}
 
\subsection{Framework Overview}
 iSmallNet consists of three streams, an interior stream, a boundary stream, and a fusion stream, as depicted in Fig. \ref{f:architecture}. 
 We first feed an infrared image to the backbone network to extract multi-scale features. 
 Then, the features of each level are fed into the interior and boundary streams supervised by decoupled labels to generate different features. 
 In the interior and boundary streams, we adopt a multi-scale nested interaction module to extract a large field of context features, and then obtain the interior features and boundary features of each stream. 
 Finally, we employ the interior-boundary fusion module in the fusion stream to integrate boundary and interior features into the overall prediction map, and generate the final map of the whole network.
 
 \textbf{1) Feature Encoder.}
 We use ResNet-18 \cite{2016-Deep-HZRS} as the backbone to extract multi-level features for the interior and boundary streams. 
 Specifically, we remove the last global pooling and full connected layers and only use five residual blocks. 
 Given an input image with the size of $H$$\times$$W$, the backbone will generate features of five scales. 
 Due to downsampling, the spatial resolution will be decreased by a stride of 2. 
 We represent these five blocks as $F$=$\left \{ F_{i} |i = 1, 2, 3, 4, 5 \right \} $. 
 The size of the feature $F_{i}$ is $\frac{W}{2^{i}}$$\times$$\frac{H}{2^{i}}$$\times$$C_{i}$, where $C_{i}$ is the number of channels. 
 In order to adapt these features for the interior/boundary map prediction, we will get two groups of features $I$=$\left \{ I_{i} |i = 1, 2, 3, 4, 5 \right \} $ and $B$=$\left \{ B_{i} |i = 1, 2, 3, 4, 5 \right \} $, which are sent to the interior and boundary streams for generating prediction maps, respectively.
 
 \textbf{2) Interior/Boundary-Stream Decoder.}
 We decouple the GT map into an interior map and a boundary one. 
 Accordingly, we develop two parallel streams for the two decoders. 
 In each stream, we use a multi-scale nested interaction module to effectively integrate the ability to detect objects of different sizes in a wide range of contexts.
 
 \textbf{3) Fusion-Stream Decoder.}
 We build a fusion-stream decoder to integrate boundary features and interior features to obtain more robust features $C$=$\left \{ C_{i} |i = 1, 2, 3, 4, 5 \right \} $ and generate the final prediction map of the whole network.
 In the fusion stream, we use an interior-boundary fusion module to integrate multi-granularity information.

\subsection{Multi-scale Nested Interaction Module}
 As shown in Fig. \ref{f:MNIM} (a), the traditional U-shaped structure is composed of an encoder, a decoder, and plain skip connections. 
 The encoder is used to expand the receptive area and extract useful information. 
 A decoder can help recover features to the same size as the input. 
 The plain skip connection acts as a bridge to transfer these low-level and high-level features from the encoder to the decoder sub-networks.
 
 Inspired by the success of nested and dense structures \cite{2018-UNet++-ZSTL} 
 and the effectiveness of the densely nested interaction module \cite{2022-DNANet-LXWWLL}, as shown in Fig. \ref{f:MNIM} (b) and (c), we design a multi-scale nested interaction module (MNIM), which overlays multiple U-shaped sub-networks to build a densely nested structure. 
 Due to the optimal receptive domains of targets with sizes varying greatly, these U-shaped sub-networks with different depths are naturally suitable for targets of different sizes. 
 Thus, we impose multiple nodes in the path between the encoder and decoder sub-networks, and all these intermediate nodes are closely connected with each other to form a nested network. 
 As shown in Fig. \ref{f:MNIM} (d), each node can receive the characteristics of its own layer and adjacent layers, resulting in repeated multi-layer feature fusion. 
 Therefore, the representation of small targets is maintained in deep layers so that better results can be obtained.
 
 As shown in Fig. \ref{f:MNIM} (d), we denote $X^{i,j}$ as the output of node $\hat{X}^{i,j}$, where $i$ is the $i^{th}$ downsampling layer along the encoder, and $j$ is the $j^{th}$ convolutional layer of a dense block along the plain skip pathway.
 When $j=0$, each node only receives the features from dense plain skip connection, and $X^{i,j}$ is formulated as:
 \begin{equation}
    X^{i,j}=F (P(X^{i-1,j})),
 \end{equation}
 where $F (\cdot )$ represents multiple cascaded convolution layers of the same convolution block, and $P (\cdot )$ means max-pooling with a stride of 2.
 
 When $j>0$, each node receives the features from the dense plain skip connection and the other four directions, including two upper layer downsampling connections, one same layer connection, and one lower layer upsampling connection, and $X^{i,j}$ is formulated as:
 \begin{equation}
 \begin{split}
    X^{i,j}=F [&P(X^{i-1,j-1}),P(X^{i-1,j}),\\
    &U(X^{i+1,j-1}),[X^{i,k}]_{k=0}^{j-1}],
 \end{split}
 \end{equation}
 where $U (\cdot )$ denotes the upsampling layer, and $[\cdot ,\cdot ]$ denotes the concatenation layer.

\subsection{Interior-Boundary Fusion Module}
 We develop a fusion module to integrate multi-granularity information. After obtaining high-quality interior and boundary prediction maps, we use a fusion stream with an interior-boundary fusion module to integrate boundary and interior features. 
 Specifically, we first extend the multi-layer features to the same size. The boundary and interior features of the same level are concatenated together. 
 Then, the shallow layer features with rich spatial and contour information and the deep layer features with rich semantic information are concatenated to generate the global robust feature maps. 
 Finally, we add the features enhanced by boundary and interior information to the global feature maps to generate the final output.

\section{Experiments}
 
\begin{table*}[!htb]\footnotesize
\centering
\caption{\label{tab:result1}mIoU, Precision, Recall and F1-score obtained by traditional, CNN-based, and our methods. Larger values in terms of the four metrics indicate higher performance. The best and second-best results are highlighted in \textcolor{red}{red} and \textcolor{blue}{blue}, respectively.}
\vspace{-2mm}
\begin{tabular}{l|c|cccc|cccc} 
\hline
\multicolumn{1}{c|}{\multirow{2}{*}{\textbf{Method}}} &\multirow{2}{*}{\textbf{Reference}}
&\multicolumn{4}{c|}{\begin{tabular}[c]{@{}c@{}}\textbf{NUAA-SIRST}
\end{tabular}} &\multicolumn{4}{c}{\begin{tabular}[c]{@{}c@{}}\textbf{NUDT-SIRST}
\end{tabular}}\\ 
\cline{3-10}
\multicolumn{1}{c|}{} & & mIoU & Precision & Recall & F1-score & mIoU & Precision  & Recall & F1-score\\ 
\hline
Filtering Based: Top-Hat \cite{1996-Detection-RJ} & OE'96         & 23.52 & 67.23 & 44.47 & 53.53 & 33.04 & 69.48 & 47.39 & 56.35 \\ 
Filtering Based: Max-Median \cite{1999-Max-DMRC}  & SPIE'99       & 18.74 & 57.35 & 60.48 & 58.87 & 20.58 & 59.47 & 56.58 & 57.99 \\ 
Local Contrast Based: LCM \cite{2013-A-CLWXT}     & TGRS'13       & 25.89 & 64.23 & 27.81 & 38.81 & 34.26 & 63.46 & 34.29 & 44.52 \\ 
Local Contrast Based: TLLCM \cite{2019-A-HMFLZ}   & TGRS'19       & 38.54 & 67.68 & 31.58 & 43.07 & 41.03 & 72.45 & 40.48 & 51.94 \\ 
Low Rank Based: IPI \cite{2013-Infrared-GMYWZH}   & TIP'13        & 33.46 & 69.29 & 64.47 & 66.79 & 30.32 & 72.49 & 60.67 & 66.06 \\ 
Low Rank Based: RIPT \cite{2017-Reweighted-DW}    & J-STARS'17    & 44.12 & 75.48 & 69.71 & 72.48 & 46.25 & 77.73 & 54.23 & 63.89 \\ 
\hline
CNN Based: TBCNet \cite{2019-TBCNet-ZCYFLW}       & arXiv'19      & 58.46 & 78.37 & 48.76 & 60.12 & 65.62 & 79.69 & 57.39 & 66.73 \\ 
CNN Based: MDvsFA-cGAN \cite{2019-Miss-WZW}       & ICCV'19       & 64.56 & 83.47 & 52.47 & 64.44 & 74.43 & 87.47 & 62.71 & 73.05 \\ 
CNN Based: ACMNet \cite{2020-Asymmetric-DWZB}     & WACV'21       & 68.96 & 87.58 & 69.61 & 77.57 & 75.41 & 89.35 & 72.56 & 80.08 \\ 
CNN Based: ALCNet \cite{2020-Attentional-DWZB}    & TGRS'21       & 72.80  & \textcolor{blue}{\textbf{88.16}} & 56.43 & 68.81 & 79.96 & 89.76 & 68.93 & 77.98 \\ 
CNN Based: DNANet \cite{2022-DNANet-LXWWLL}       & TIP'22        & \textcolor{blue}{\textbf{77.47}} & 83.37 & \textcolor{blue}{\textbf{86.23}} & \textcolor{blue}{\textbf{84.78}} & \textcolor{blue}{\textbf{86.53}} & \textcolor{red}{\textbf{91.98}}  & \textcolor{blue}{\textbf{92.76}} & \textcolor{blue}{\textbf{92.37}}  \\ 
\hline
\textbf{iSmallNet} (full model) & \textbf{Ours} & \textcolor{red}{\textbf{80.34}}  & \textcolor{red}{\textbf{88.74}}  & \textcolor{red}{\textbf{90.79}}  & \textcolor{red}{\textbf{89.75}}  & \textcolor{red}{\textbf{87.25}}  & \textcolor{blue}{\textbf{90.26}} & \textcolor{red}{\textbf{95.55}}  & \textcolor{red}{\textbf{92.83}}   \\
\hline
iSmallNet w/o interior/boundary stream      & Ablation & 79.24 & 86.96 & 88.67 & 87.81 & 85.67 & 88.56 & 93.89 & 91.15 \\
iSmallNet-UNet (w/o MNIM)                   & Ablation & 78.74 & 86.35 & 88.13 & 87.23 & 85.13 & 88.38 & 93.74 & 90.98 \\
iSmallNet-UNet++ (w/o MNIM)                 & Ablation & 79.09 & 87.14 & 88.78 & 87.95 & 86.19 & 88.69 & 94.09 & 91.31 \\
iSmallNet-DNANet (w/o MNIM)                 & Ablation & 79.26 & 87.41 & 89.46 & 88.42 & 86.56 & 89.63 & 94.81 & 92.15 \\ 

\hline
\end{tabular}
\end{table*}

 \subsection{Datasets and Implementation}
 We conduct experiments on two popular benchmarks, \emph{i.e.}, NUAA-SIRST \cite{2020-Asymmetric-DWZB} and NUDT-SIRST \cite{2022-DNANet-LXWWLL}. NUAA-SIRST is currently the largest realistic ISTD dataset, including clouds, sea surface, buildings, \emph{etc}. 
 The scales of the targets are varied, the large-scale targets are close to more than 20 pixels, while the small-scale targets only occupy 1 to 2 pixels. NUDT-SIRST consists of five main scenes (\emph{i.e.}, cities, fields, highlights, seas and clouds) and introduces more challenging scenes (\emph{e.g.}, multiple targets, point targets, and dim target scenes). 
 Each image is synthesized with different targets under various signal-to-clutter ratios (SCR) and rich poses.
 
 We train the whole network by using stochastic gradient descent (SGD) with a momentum of 0.9 and a weight decay of 0.0005. 
 The batch size is set to 16, and the epoch is set to 1500. 
 The initial learning rate is set to 0.05, and the strategy of learning rate preheating and linear attenuation is adopted. 
 We adopt widely-used metrics, including mIoU, precision, recall, and F1-score, to evaluate all methods.

 \subsection{Comparison with State-of-the-Art Methods}
\begin{figure}[t]
	\centering
	\includegraphics[width=1\linewidth]{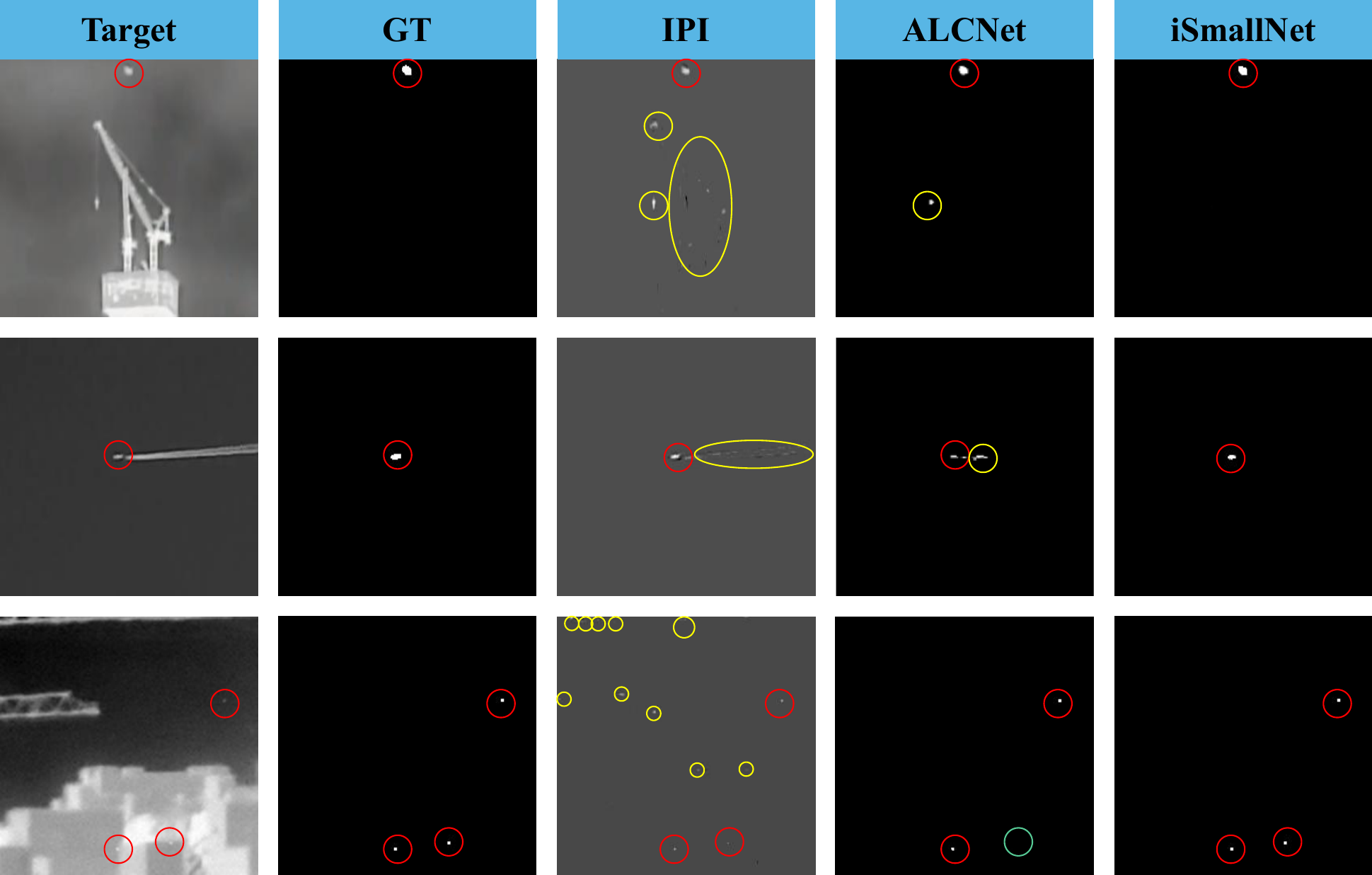}
	\caption{Qualitative results. The correctly detected target, false alarm, and missed detection areas are highlighted by \textcolor{red}{red}, \textcolor{yellow}{yellow}, and \textcolor{green}{green} dotted circles, respectively.}
	\label{f:predict} \vspace{-3mm}
 \end{figure}

 We compare our iSmallNet to several state-of-the-art methods, including traditional methods
 (Top-Hat \cite{1996-Detection-RJ},
 Max-Median \cite{1999-Max-DMRC},
 LCM \cite{2013-A-CLWXT},
 TLLCM \cite{2019-A-HMFLZ},
 IPI \cite{2013-Infrared-GMYWZH},
 and RIPT \cite{2017-Reweighted-DW}),
 and CNN-based methods
 (TBCNet \cite{2019-TBCNet-ZCYFLW},
 MDvsFA-cGAN \cite{2019-Miss-WZW},
 ACMNet \cite{2020-Asymmetric-DWZB},
 ALCNet \cite{2020-Attentional-DWZB},
 and DNANet \cite{2022-DNANet-LXWWLL}).\\
 \textbf{1) Quantitative Results.}
 As depicted in Table \ref{tab:result1}, traditional methods are more suitable for specific scenes, and CNN-based methods are relatively better than traditional methods. 
 Compared with both traditional and CNN-based methods, our iSmallNet can indeed achieve significant improvement, especially on the NUAA-SIRST dataset. 
 This is because the two datasets contain different cluttered backgrounds, and the shape and size of small targets vary largely, and our interior and boundary streams can effectively learn the interior and boundary information of small targets. 
 Besides, we further learn enhanced representations of small targets through the proposed MNIM, so the inherent characteristics of infrared small targets can be fully learned in our iSmallNet.\\
 \textbf{2) Qualitative Results.}
 As demonstrated in Fig. \ref{f:predict}, traditional methods are prone to false alarms, while CNN-based methods are relatively easy to miss detection of dim-small targets. 
 Our iSmallNet can produce accurate target location and shape segmentation outputs at a very low false alarm rate.
 
 \subsection{Ablation Study}
 We demonstrate the comparative results of iSmallNet and its variants in the last 5 rows of Table \ref{tab:result1}. 
 We can see that both interior and boundary streams can effectively advance the detection performance. 
 Furthermore, iSmallNet achieves significant improvement by adding the MNIM.

\section{Conclusion}
 In this paper, we propose a novel label decoupling-based densely nested network, namely iSmallNet, for ISTD. Different from current CNN-based efforts, iSmallNet adopts a three-stream network to leverage the interior and boundary information for better detecting small targets. To make full use of decoupling supervisions and maintain small targets in deep layers, we design a multi-scale nested interaction module to explore a wide range of context information. Moreover, an interior-boundary fusion module is developed to integrate multi-granularity information to further enhance the detection capability. Experiments on two benchmarks verify that iSmallNet performs favorably against state-of-the-art detectors.
 
\noindent\textbf{Acknowledgments.} 
 This work was partially supported by the National Natural Science Foundation of China (No. 62276129) and the Natural Science Foundation of Jiangsu Province (No. BK20220890).

\bibliographystyle{IEEEbib}
\bibliography{ref}

\end{document}